\pdfoutput=1

\documentclass[11pt]{article}

\usepackage{acl}
\usepackage{amsmath} 
\usepackage{times}
\usepackage{latexsym}
\usepackage[ruled]{algorithm2e}
\usepackage{tabularx}
\usepackage[T1]{fontenc}

\usepackage[utf8]{inputenc}
\usepackage{enumitem}
\usepackage{microtype}

\usepackage{inconsolata}

\usepackage{graphicx}
\title{FastLexRank: Efficient Lexical Ranking for Structuring Social Media Posts}

\author{ Mao Li\\
  Institute for Social Research \\
  University of Michigan\\
  Ann Arbor, MI 48104 \\
  \texttt{maolee@umich.edu} \\
\And
Frederick Conrad\\
  Institute for Social Research \\
  University of Michigan\\
  Ann Arbor, MI 48104 \\
  \texttt{fconrad@umich.edu} \\
\And
Johann Gagnon-Bartsch\\
  Department of Statistics \\
  323 West Hall\\
  Ann Arbor, MI 48104 \\
  \texttt{johanngb@umich.edu}
}

\begin{document}
\maketitle
\begin{abstract}
We present FastLexRank\footnote{https://github.com/LiMaoUM/FastLexRank}, an efficient and scalable implementation of the LexRank algorithm for text ranking. Designed to address the computational and memory complexities of the original LexRank method, FastLexRank significantly reduces time and memory requirements from $\mathcal{O}(n^2)$ to $\mathcal{O}(n)$ without compromising the quality or accuracy of the results. By employing an optimized approach to calculating the stationary distribution of sentence graphs, FastLexRank maintains an identical results with the original LexRank scores while enhancing computational efficiency. This paper details the algorithmic improvements that enable the processing of large datasets, such as social media corpora, in real-time. Empirical results demonstrate its effectiveness, and we propose its use in identifying central tweets, which can be further analyzed using advanced NLP techniques. FastLexRank offers a scalable solution for text centrality calculation, addressing the growing need for efficient processing of digital content.
\end{abstract}

\section{Introduction}
In recent years, social media has emerged as a crucial data source for public opinion research \citep{murphy2014social, mneimneh2021data, jensen2021analysis}. Beyond traditional methodologies such as sentiment analysis and topic modeling, text summarization methods have gained prominence for distilling the essence of discussions within the social media landscape. Text summarization is an NLP task designed to condense a set of documents into a succinct representation of their gist \cite{abd_elaziz_text_2020}. There are two main approaches to text summarization: abstractive and extractive \cite{abd_elaziz_text_2020}. The abstractive method involves rephrasing the original text in shorter human-like narratives, abstracting away from the details, whereas the extractive method involves selecting specific sentences from the original documents that encapsulate the key ideas \cite{erkan_lexrank_2004}.

These approaches enable researchers to capture and comprehend the vast array of conversations and viewpoints expressed online, providing insights into public sentiments and trends. Abstractive and extractive text summarization techniques have been employed in various applications, such as real-time event detection on Twitter using extractive methods \cite{alsaedi_automatic_2021}, summarizing opinionated texts \cite{liang_research_2012}, and identifying "event messages" within large volumes of tweets \cite{becker_selecting_2021}. Other studies have applied the abstractive approach to news reports and social media data, such as Reddit posts \cite{kim2018abstractive, zhan2022you}, and Twitter/X \cite{blekanov2022transformer, li2020abstractive}.

Recent developments in Large Language Models (LLMs) have substantially enhanced the capabilities of abstractive summarization, yielding outputs of notable quality. Models such as PEGASUS \cite{zhang2020pegasus} and advanced LLMs like ChatGPT and Llama2 \cite{touvron2023llama} have demonstrated exceptional proficiency in condensing extensive texts into coherent and concise summaries. However, the fixed context windows of transformer-based models constrain their ability to process and distill exceedingly large text corpora. Although innovations like Gemini 1.5 Pro \citep{team2023gemini}\footnote{https://blog.google/technology/ai/google-gemini-next-generation-model-february-2024}, with a context window of 1 million tokens, are now operational, the computational load remains substantial for very lengthy documents due to the self-attention mechanism's complexity, denoted by $\mathcal{O}(n^2 \cdot d)$. This complexity underscores the challenges in scaling summarization tasks for extensive texts without incurring significant computational costs, explaining why previous summarization studies mostly focus on single documents/posts/threads rather than the entire relevant corpus.

Therefore, the question becomes, when facing millions of social media posts, how can we quickly identify the most important and representative posts to distill their information? Similar to the idea from Retrieval Augmented Generation (RAG) \citep{lewis2020retrieval}, can we first target the most relevant posts and then ask LLMs to generate a summary based on this content? However, current retrieval models rely heavily on correctly specified queries to perform nearest-neighbor searches and cannot self-rank the posts based on their centrality or representativeness. This is where traditional extractive text summarization methods, like LexRank, can augment text summarization with ordering.

LexRank, introduced by \citet{erkan_lexrank_2004}, applies the principles of the PageRank algorithm \citep{BRIN1998107} to a graph representation of sentences, calculating the importance of each sentence within the corpus. It uses TF-IDF representations to construct a graph where the nodesrepresent sentences.. As an automatic summarization technique, LexRank has shown a remarkable ability to identify the most salient texts (with high centrality scores) within a set of documents. Despite its limitations in coherence and consistency inherent to extractive approaches, LexRank's ability to pinpoint the most representative text segments is invaluable in data mining and information retrieval. Unlike transformer-based language models, LexRank can theoretically analyze texts of unlimited length, using centrality scores to determine their typicality. This feature is particularly advantageous for distilling core information and bringing order to social media posts from large-scale textual datasets.

In this study, we reinterpret and expand the traditional scope of the LexRank algorithm beyond its original function as merely a text summarization tool, proposing its application as a comprehensive ranking algorithm. We do not treat LexRank as an alternative to LLM summarization; instead, we propose it as an augmented method that can help LLMs address context window limitations and make the summarization process more efficient.

By conceptualizing the selected social media sample as an extended document and viewing each tweet as an individual sentence, LexRank can effectively prioritize tweets based on their centrality or representativeness, highlighting how each tweet relates to the corpus as a whole. Consequently, this approach enables the identification of a subset of tweets that most accurately represents the corpus. In public opinion research, showcasing actual tweets rather than AI-generated paraphrases becomes important and informative. In this context, an extractive method—or, more aptly, a lexical ranking algorithm—offers significant contributions to leveraging social media for public opinion analysis.

Despite requiring fewer computational resources than LLMs, the immense volume of data in social media corpora continues to present a substantial challenge to the original LexRank approach, given its $\mathcal{O}(n^2)$ time and space complexity. For LexRank to remain effective and relevant, it must be adapted to handle and process large-scale data efficiently. This adaptation is essential for extracting meaningful and representative summaries from extensive and continuously expanding digital content, which is our test domain.

Accordingly, our research introduces FastLexRank, a novel approach to improving the efficiency of ranking texts using LexRank. By leveraging LexRank for its ranking capabilities, we can organize the massive volume of social media posts based on their centrality scores. These scores enable us to identify and select the most representative posts, which LLMs can then summarize into a coherent and comprehensive narrative. This improvement not only addresses the computational challenges posed by large datasets but also enhances the applicability of LLMs in generating concise summaries to voluminous data.
\begin{algorithm}[ht]
\caption{Streamlined LexRank Algorithm for Centrality Scores}
\label{al1}
\KwIn{Corpus $C$}
\KwOut{Centrality Scores Vector $Scores$}

$Sentences \leftarrow \text{ExtractSentences}(C);$

$Embedding \leftarrow \text{Embeddings}(Sentences);$

$SimMat \leftarrow \text{ComputeSim}(Embedding);$

$TransMat \leftarrow \text{Transition}(SimMat);$

$Scores \leftarrow \text{PowerMethod}(TransMat);$

\normalsize
\Return{$Scores$};

\end{algorithm}

\section{Limitation of Original LexRank}

The streamlined original implementation of the LexRank algorithm is outlined in Algorithm~\ref{al1}. It is important to note that in this study, we have not taken into account additional hyperparameters, such as the threshold for similarity scores and the damping factor; instead, we assume a fully connected corpus graph. As demonstrated, although LexRank proves effective for text summarization, its space, and computational complexities present considerable challenges when applied to vast datasets, including those with millions of posts.

\begin{algorithm}[ht]

\SetAlgoLined
\KwIn{Sentence embedding matrix \(E\), tolerance \(\epsilon\), maximum iterations \(max\_iter\)}
\KwOut{Centrality scores \(c\)}
\BlankLine
\tcp{Compute the similarity matrix \(S\)}
\For{\(i = 1\) \KwTo \(n\)}{
    \For{\(j = 1\) \KwTo \(n\)}{
        \(S[i, j] \leftarrow \frac{E[i] \cdot E[j]}{|E[i]| \cdot |E[j]|}\);
    }
}

\tcp{Derive the transition matrix \(M\) from \(S\)}
\For{\(i = 1\) \KwTo \(n\)}{
    \(M[i] \leftarrow \frac{S[i]}{\sum S[i]}\);
}

\(c \leftarrow \text{random vector of length } n\);

\(c \leftarrow \frac{c}{|c|_1}\);

\For{\(i = 1\) \KwTo \(max\_iter\)}{
    \(c_{old} \leftarrow c\);

    \(c \leftarrow M^T \cdot c\);

    \(c \leftarrow \frac{c}{|c|_{1}}\);

    \If{\(|c - c_{old}|_1 < \epsilon\)}{
        \tcp{convergence reached}
        \KwRet{\(c\)};
    }
}

\KwRet{\(c\)};
\caption{Power Method for Computing LexRank scores}
\label{old_power_method}
\end{algorithm}

A significant limitation of LexRank pertains to its memory complexity, quantified as $\mathcal{O}(n^2)$, attributable to the requirement of forming a dense stochastic matrix. Shifting the focus to computational complexity, another notable challenge emerges. The primary computational burden stems from calculating the stationary distribution of the Markov chain, which requires determining the eigenvector associated with the eigenvalue of one. In this context, the power method, as detailed in the original LexRank paper, becomes crucial. The specifics of this method are further expounded in Algorithm~\ref{old_power_method}.

In conclusion, the computational complexity of the power method, integral to LexRank, is $\mathcal{O}(n^2)$. This complexity primarily arises from the matrix-vector multiplication in each iteration, involving the sentence similarity matrix of size $n \times n$. As $n$, the number of sentences, increases, the computational demands escalate, posing a major bottleneck in the efficiency of the LexRank algorithm when dealing with large-scale documents.

\section{FastLexRank Approach}

We consider embeddings where, without loss of generality, it is assumed that each embedding vector is normalized to unit length, i.e., \(|E[i]| = 1\). Consequently, \(S = EE^T\), where $E$ represents the embedded vector for each document (i.e., the posts in the social media analysis scenario), and $S$ represents the covariance matrix of embedding vectors. Let \(\sigma \equiv S\mathbf{1}\) represent the row sums of \(S\). Further, let \(\Sigma\) denote the diagonal matrix with diagonal entries corresponding to \(\sigma\). Thus, \(M = \Sigma^{-1}S\), and centrality scores are obtained by identifying the eigenvector of \(M^T\) corresponding to the eigenvalue 1. It is important to note that \(\sigma\) is this eigenvector, since \(M^T\sigma = EE^T\Sigma^{-1}\sigma = EE^T\mathbf{1} = \sigma\).

Hence, in theory, centrality scores can be computed solely using the text embedding matrix \(E\). Initially, calculating \(E^T\mathbf{1}\) yields the column sum vector \(z\) of \(E\). Subsequent matrix multiplication \(Ez\) then produces the centrality score, achievable within linear time complexity. This approach is not only mathematically straightforward but also computationally simpler compared to the traditional power method, offering identical results.

It is also, we feel, more interpretable.  Firstly, note that $E^T\mathbf{1} = \frac{1}{n}\bar{E}$, where $n$ is the number of sentences (or tweets in our case) in the corpus, and $\bar{E}$ is the mean embedding.  Thus, $EE'\mathbf{1} \propto E\bar{E} \propto S_C (E, \bar{E})$, where $S_C$ denotes cosine similarity.  In other words, the typicality score of any given sentence embedding is simply (after rescaling) the cosine similarity of that embedding with the corpus' mean embedding --- i.e., how similar that sentence is to the average sentence.

The pseudo-code demonstrating this method is presented in Algorithm~\ref{fast}.

\begin{algorithm}[ht]
\caption{FastLexRank Method}
\label{fast}
\KwIn{Sentence Embeddings matrix $E$}
\KwOut{ Centrality Score $c_a$}


$z \leftarrow \text{ColumnSum}(E)$\;

\For{$i = 1$ \KwTo $\text{length}(z)$}{
  $z[i] \leftarrow \frac{z[i]}{\sqrt{\sum(z[i]^2)}}$\;
}

$c_a \leftarrow E \cdot z$\;

\Return{$c_a$}\;
\end{algorithm}

\section{Experiment}
Even though we have formally proven that our approach derives the same LexRank centrality ranking as the original LexRank through theoretical analysis, it is still necessary to empirically validate this finding. Therefore, the experimental questions that guide our work are as follows:
\begin{enumerate}
    \item Equivalence of Results: Does our implementation of FastLexRank yield identical centrality ranking compared to the original LexRank algorithm across a variety of datasets?
    
    \item Practical Efficiency Gains: How does FastLexRank perform in practice in terms of computational speed, particularly when applied to large, real-world datasets like social media corpora? Specifically, how significant are the efficiency gains when processing large-scale data compared to the original LexRank?
\end{enumerate}

\subsection{Dataset}

As we discussed before, the FastLexRank algorithm proves particularly useful for ranking text centrality in large-scale corpora, such as posts from social media. To evaluate the performance of FastLexRank against the original LexRank algorithm, we conducted experiments using a Twitter corpus focused on U.S. political discussions, comprising 2004 tweets \cite{marchetti2012learning}. This dataset serves as an ideal testbed to assess LexRank's efficacy in identifying key discussions or micro-blogs (tweets) within a substantial corpus.

\subsection{Experimental Setting}
This study contrasts our methodology, which calculates a centrality score, with the conventional power method. Specifically, we utilized the degree\_centrality\_scores function from the Python \href{https://pypi.org/project/lexrank/}{lexrank package} to implement the traditional power method. A significant aspect of this comparison is our integration of a novel sentence embedding technique alongside the traditional \textbf{TF-IDF} representation, allowing for a comprehensive evaluation of our approach across various word representation methods. Two text representations were constructed using \texttt{TfidfVectorizer} from the \href{https://scikit-learn.org/stable/modules/generated/sklearn.feature_extraction.text.TfidfVectorizer.html}{scikit-learn package}, and \texttt{SentenceTransformer} class with ``all-MiniLM-L6-v2'' model from the \href{https://www.sbert.net/}{sentence-transformer package}. This comparison elucidated the differences in speed and performance between the two algorithms.

Also, we want to note that our experiments were carried out on a high-performance computing cluster, configured with Redhat8, Intel Xeon Gold 6226R CPUs, 180 GB of RAM, and 1 NVIDIA A40 GPU, with 48 GB of VRAM.

\subsection{Results}
In the evaluation of LexRank as a ranking algorithm, a pivotal factor to consider is the consistency of the outcomes it yields across various implementations. This study focuses on whether different implementations of LexRank (i.e., our FastLexRank and original LexRank) yield closely aligned centrality scores and ranking sequences, a criterion for considering them as effective improvement. We present a scatter plot comparing vectors of centrality scores with the scores computed using the original LexRank method. Furthermore, we augment this analysis with Kendall's tau (\(\tau\)) test, conducted using the \texttt{kendalltau} function from the SciPy package (see \href{https://docs.scipy.org/doc/scipy-0.15.1/reference/generated/scipy.stats.kendalltau.html}{SciPy documentation}). The computation of Kendall's \(\tau\) is shown in Equation~\ref{eq1}, where \(P\) denotes the number of concordant pairs, \(Q\) the number of discordant pairs, \(T\) the number of ties in \(x\) only, and \(U\) the number of ties in \(y\) only. Pairs tied in both \(x\) and \(y\) are excluded from \(T\) and \(U\)\footnote{\url{https://docs.scipy.org/doc/scipy-0.15.1/reference/generated/scipy.stats.kendalltau.html}}. This quantitative analysis compares the ranked sequences generated by both the original LexRank and the FastLexRank approach, providing a rigorous evaluation of their alignment. Values of \(\tau\) near 1 suggest strong agreement, whereas values near -1 indicate strong disagreement.

\begin{equation}
    \tau = \frac{P - Q}{\sqrt{(P + Q + T) \times (P + Q + U)}}
\label{eq1}
\end{equation}

\begin{figure}
\centering 
\includegraphics[width=0.5\textwidth]{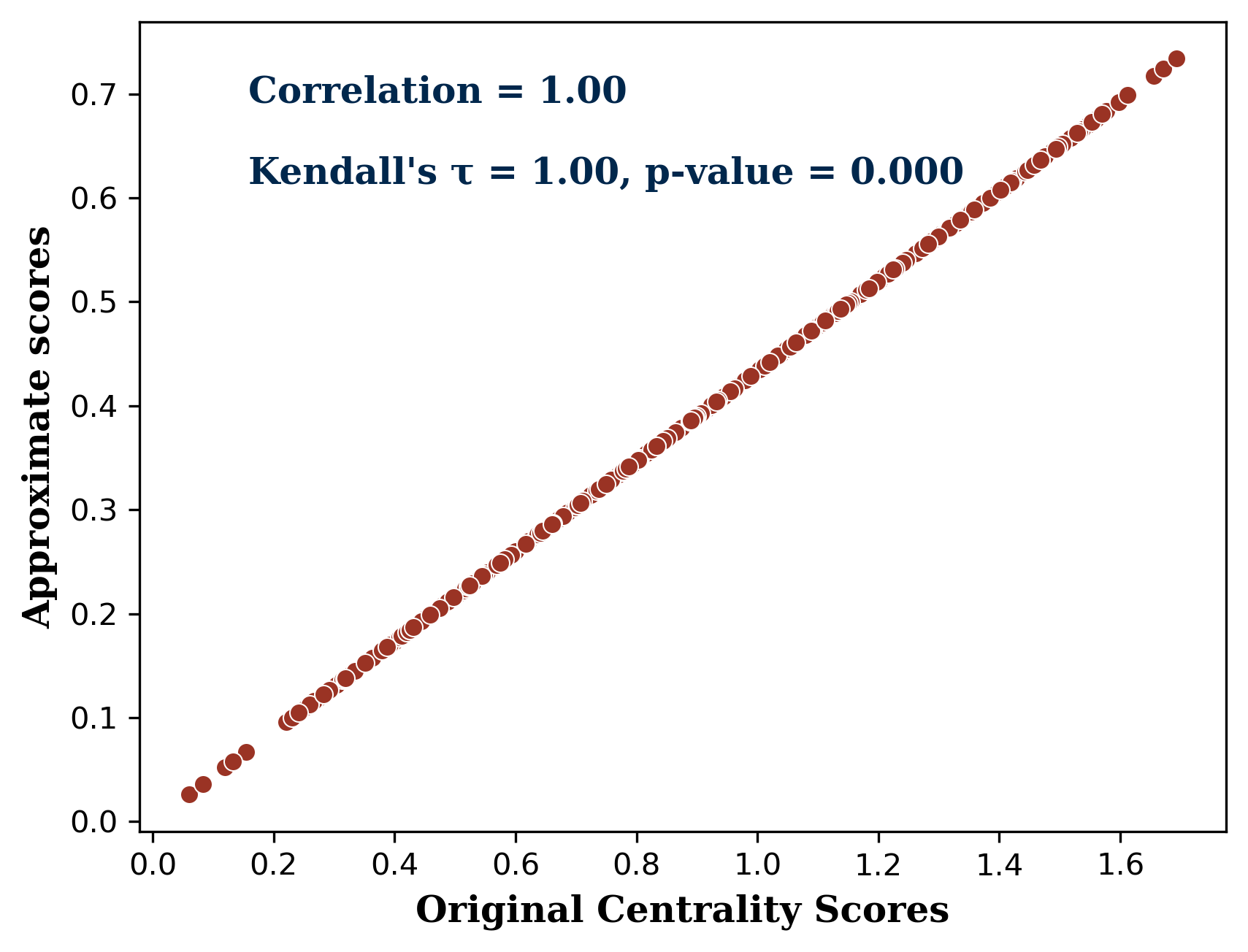} 
\caption{Comparison of FastLexRank Centrality Scores and Original Centrality Scores Using SBERT Embeddings} 
\label{fig:sbertC}
\end{figure}

Figure~\ref{fig:sbertC} depicts the scatter plot between FastLexRank centrality scores and original centrality scores utilizing \textbf{SBERT} embeddings. The alignment of the two vectors of scores is striking, with the correlation coefficient reaching 1.0, indicating a perfect positive linear correlation. This close correlation is a strong indication of the reliability of the FastLexRank method. Further, we assessed the consistency of the ranking sequences derived from these two approaches. The Kendall's $\tau$ statistics is 1, indicating the rankings obtained using the FastLexRank and the original algorithm were found to be identical, i.e., there are no discordant pairs. The results suggest no difference (at all) in terms of ranking efficacy between the two methods.

\begin{figure}
    \centering
    \includegraphics[width=0.5\textwidth]{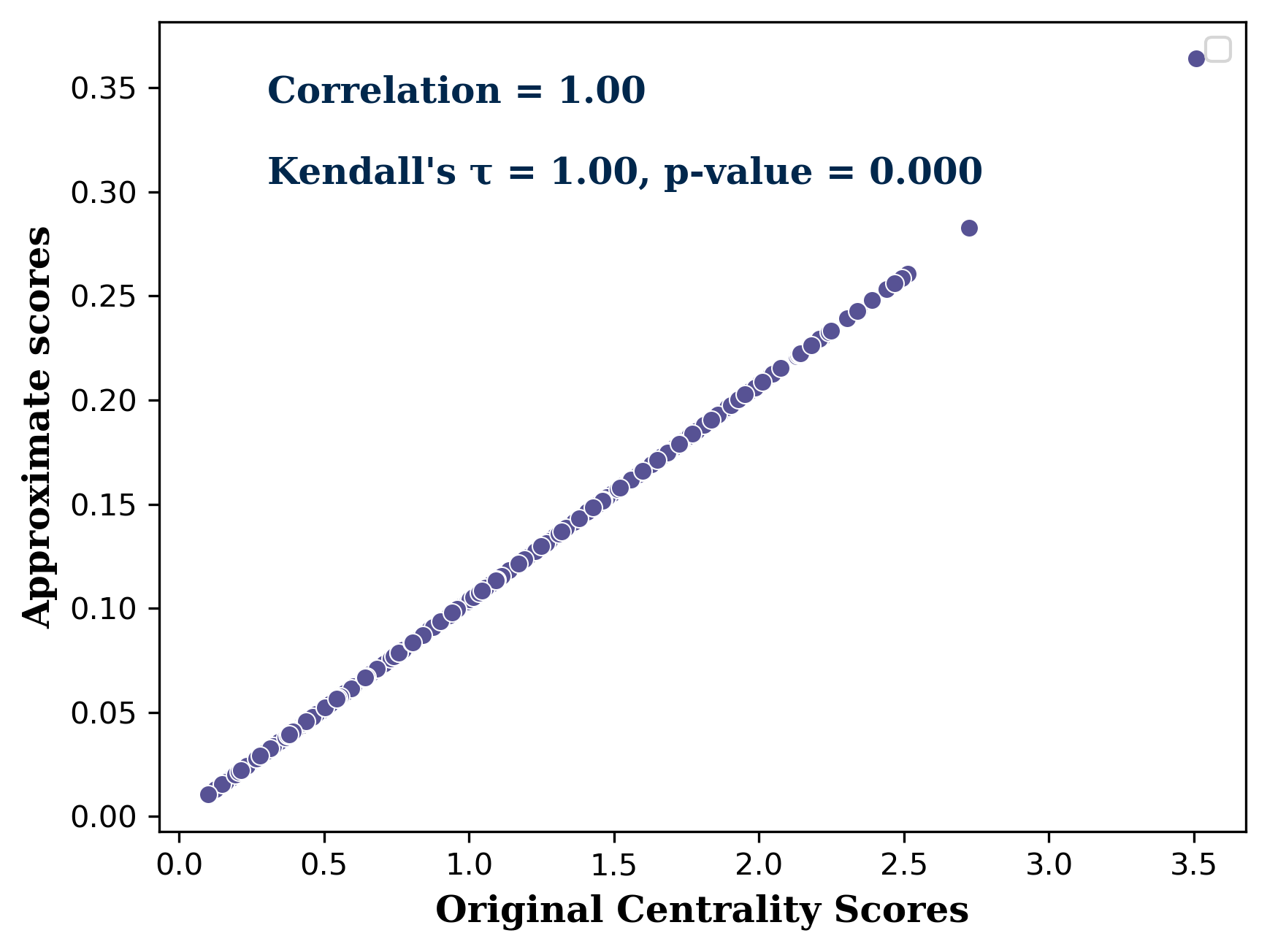}
    \caption{Comparison of FastLexRank Scores and Original Centrality Scores Using TF-IDF Embeddings}
    \label{fig:tfidfC}
\end{figure}

Furthermore, we conducted a similar comparison analysis using \textbf{TF-IDF} embeddings. In this study, we applied stop-word filtering and did not filter the term by minimum frequency while constructing the \textbf{TF-IDF} representation. Figure~\ref{fig:tfidfC} presents results identical to those obtained using \textbf{SBERT} embeddings, i.e., the correlation of the scores is 1.0, and the ranking sequence is identical. The perfect correlation and identical ranking sequence indicate that the FastLexRank algorithm performs equally well in the \textbf{TF-IDF} representation. When ranking tweets, the resulting sequences are identical, whether using the power method or our approach.

\begin{figure}
    \centering
    \includegraphics[width=0.5\textwidth]{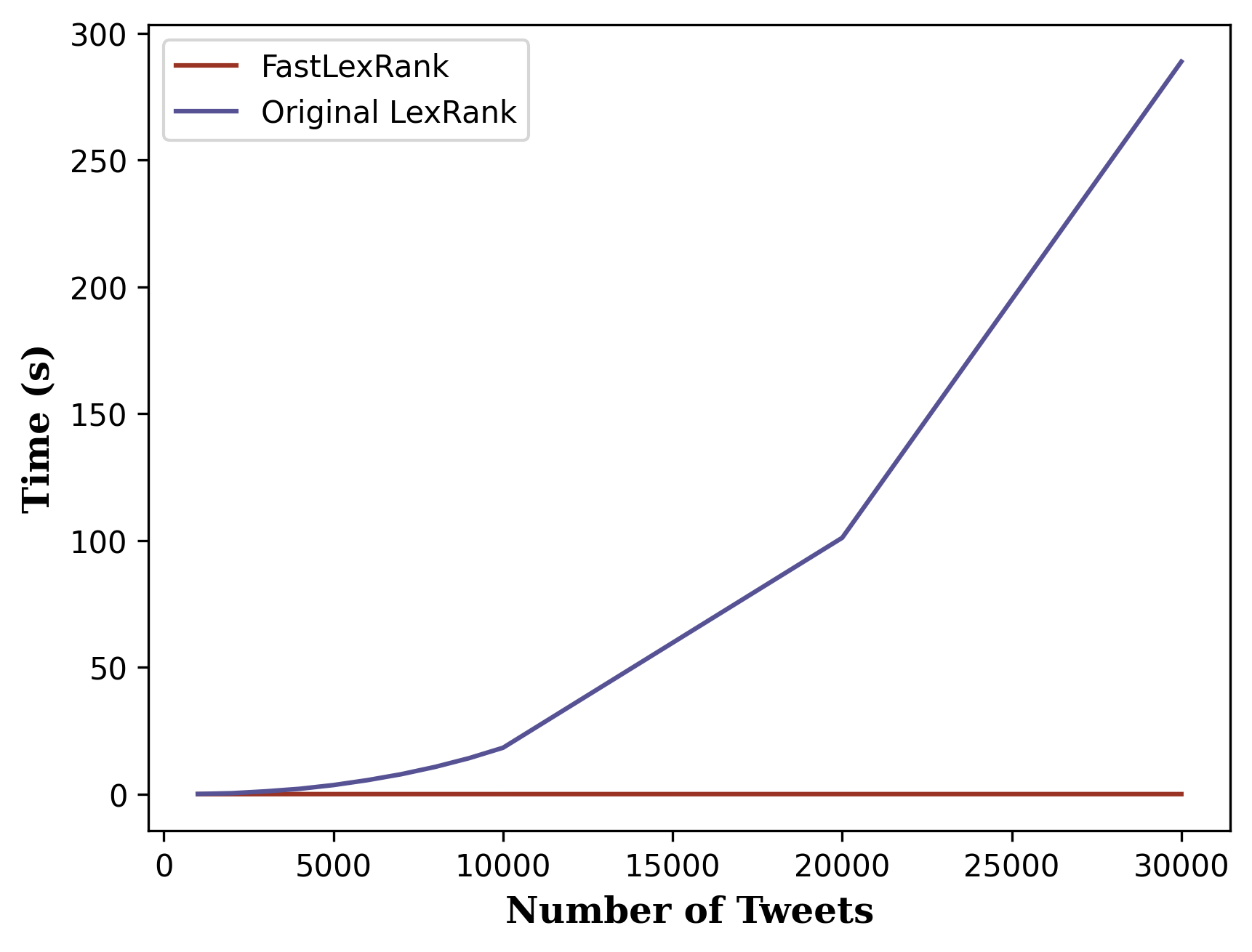}
    \caption{Time spent on Fast LexRank and Original LexRank}
    \label{fig3}
\end{figure}

\subsection{Assessment of Robustness}

\begin{table*}[ht]
\resizebox{2.1\columnwidth}{!}{%
    \centering
    \begin{tabular}{|c|c|c|c|c|c|c|c|c|c|c|c|c|}
    \hline
       Size & 1,000 & 2,000 & 3,000 & 4,000 & 5,000 & 6,000 & 7,000 & 8,000 & 9,000 & 10,000 & 20,000 & 30,000 \\
       \hline
        Kendall's $\tau$ & 1.0 & 1.0 & 1.0 & 1.0 & 1.0 & 1.0 & 1.0 & 1.0 & 1.0 & 1.0 & 1.0 & 1.0  \\ \hline
        Correlation & 1.0 & 1.0 & 1.0 & 1.0 & 1.0 & 1.0 & 1.0 & 1.0 & 1.0 & 1.0 & 1.0 & 1.0 \\ \hline
    \end{tabular}
    }
    \caption{Assessment of Robustness on different size of corpora}
    \label{table2}
\end{table*}

To assess the robustness of our method, we expanded our evaluation to include additional datasets, particularly selecting various Twitter corpora that contain keywords pertinent to the U.S. 2020 Census, spanning from January 1st, 2020, to February 29th, 2020. This corpus comprises 189,496 unique tweets. We randomly selected 12 subsets from the original corpus, with sizes ranging from 1,000 to 30,000 tweets. In addition to evaluating the correlation of scores and Kendall's $\tau$, we also measured the computational time of each method, excluding the time required for embedding phrases, as it is a prerequisite step for both our method and the original LexRank algorithm. The outcomes of these evaluations are detailed in Table~\ref{table2}, and the computational times are illustrated in Figure~\ref{fig3}.

As depicted in Table~\ref{table2}, our algorithm exhibits consistent performance across the diverse social media datasets. While our FastLexRank algorithm does not always yield identical ranking results as the original algorithms, the variations are slight, highlighting the efficacy of our approach.\footnote{The reason for not achieving identical results lies in the use of the power method, which is an approximation algorithm. The iterative nature of the power method can introduce small errors, resulting in minor differences in ranking outcomes.}

Figure~\ref{fig3} shows that while computing time for the original LexRank algorithm increases sharply the more tweets it is processing, there is no increase in computing time for FastLexRank. This radically improved performance for FastLexRank is anticipated as the algorithm's time complexity of $\mathcal{O}(n)$ in contrast to the $\mathcal{O}(n^2)$ time complexity of the power method employed by the original LexRank for computing the stationary distribution. 

\section{Case Study}
In this section, we demonstrate how to integrate FastLexRank with other Natural Language Processing (NLP) pipelines, such as text summarization. Using the citizenship dataset, we collected a total of 189,496 unique tweets. Utilizing the OpenAI tokenizer toolkit, \href{https://github.com/openai/tiktoken}{tiktoken}, we identified 9,805,237 tokens that need processing. Summarizing this entire corpus with GPT-4o would encounter context window limitations. While breaking the corpus into digestible chunks for GPT-4o is possible, the coherence of the overall summarization would be challenging to maintain. Moreover, the cost of using the OpenAI API (for input alone) would be approximately \$50.

We also observe that keyword searches within social media corpora often yield noisy posts, which may not align with the target discussion. Incorporating such noise into the LLM-generated summaries would dilute the quality of the information distilled. To highlight this, Table~\ref{tab:mlRepre} presents the two most representative and two least representative posts, as determined by the FastLexRank similarity score. This table underscores the necessity of filtering social media corpora to focus on posts containing topical information.

\begin{table*}[]
    \centering
    \begin{tabular}{p{2cm}|p{12cm}}

    \hline
     Centrality    & Tweet   \\
     \hline
     0.87 & \#Census2020 is so important for our community because \#WeCount. It will determine:School funding, Roads, public transportation, and infrastructure, Community Resources, Political representation …for the next 10years!\\
     \hline
     0.87 & It's important to our future. The Census 2020 is coming. Get more information at 2020census.gov.\\
     
     \hline\hline
    -0.06   &  @uscensusbureau Beside the verse dramas, with their lyrical interludes, on which many operas were based, there were poetic texts which were set as secular cantatas. One of the earliest was Alessandro Stradella's La Circe, in a setting for three voices that bordered on the operatic.\\
    \hline
     -0.11   & Which can mislead you one key ways to know the difference is to always trust your gut feelings focus on the sensations you feel in your body at any given situation because what your soul sees your body always feels you're so census and reads energy and vibrations, cause it is  \\
     \hline
    \end{tabular}
    \caption{Most and Least Representative posts Identified by FastLexRank}
    \label{tab:mlRepre}
\end{table*}

Leveraging FastLexRank streamlines the summarization pipeline for the 189,496 tweets. Initially, we apply the FastLexRank algorithm to the original corpus, selecting the top 100 most representative posts. We then summarize these posts using LLMs. This approach yielded a comprehensive summary from GPT-4o, covering seven distinct topics related to Census 2020. The complete GPT-4o response is available in Appendix~\ref{sec:appendix}.

\section{Discussion}
Our study reveals that the implementation of our proposed method markedly decreases the time complexity of computing centrality scores, from $\mathcal{O}(n^2)$ to $\mathcal{O}(n)$. Furthermore, by replacing the conventional approach for determining the stationary distribution with our technique, we effectively reduce the overall time complexity of the LexRank algorithm from $\mathcal{O}(n^2)$ to $\mathcal{O}(n)$. In this revised framework, the majority of the computational time is allocated to constructing the sentence embedding matrix. Additionally, our method also reduces the memory requirements during the calculation process.

The FastLexRank algorithm introduces a novel mechanism for the rapid assessment of text centrality or representativeness within expansive text corpora. This utility is especially pronounced in two primary use cases. Firstly, it facilitates the expedited extraction of salient information from large-scale corpora, such as social media datasets, enabling researchers to swiftly pinpoint central tweets encapsulating the corpus's overarching narrative. This rapid identification of central messages significantly streamlines the initial phase of qualitative analysis, allowing for immediate insights into the corpus content.

Moreover, the computational efficiency of FastLexRank permits the preliminary selection of the top \(n\) central posts within a voluminous dataset. Subsequently, integrating these identified posts with LLMs yields enriched, coherent summaries. This methodology substantially enhances the capacity of social media researchers to efficiently navigate and interpret extensive datasets, thereby broadening their understanding of prevalent user discourses. Furthermore, this approach is amenable to integration with other NLP methodologies, such as topic modeling, to delineate dominant conversations within each thematic cluster, thereby augmenting the analytical granularity afforded to researchers in the domain of computational linguistics.
 
We also believe this method can inform other ranking algorithms inspired by PageRank or similar graph-based algorithms. In summary, the rapidly evolving information landscape necessitates more efficient methods to handle the vast amount of information generated daily.

\section{Limitation}

A key assumption in FastLexRank's design is treating sentence graphs as fully connected. This approach simplifies the computational model, facilitating significant efficiency gains. However, it inherently overlooks the potential benefits of thresholding—a process that could mitigate noise by disregarding connections between sentences with negligible similarity. Implementing such a threshold might filter out irrelevant connections, but it comes at the cost of reducing the model's flexibility and potentially overlooking subtle yet relevant relationships.

 FastLexRank significantly advances the efficiency of the original LexRank algorithm, a foundational technique in lexical ranking that has stood for decades. By integrating the idea of self-ranking, we aim to revitalize this classical approach. Nonetheless, the reliance on LexRank's fundamental principles means that FastLexRank may inherit some of its predecessor's limitations, particularly in adapting to the nuances of modern data sets and evolving linguistic patterns.

FastLexRank represents a significant advance in modeling efficiency. However, it is crucial to recognize the evolving landscape of retrieval models, particularly those leveraging transformer-based deep learning architectures. These cutting-edge algorithms have demonstrated potential superiority in identifying the most 'representative' elements within datasets, including tweets. Although the majority of contemporary models prioritize query-based retrieval, our acknowledgment of these advances underscores an essential direction for future research. Specifically, there is a pressing need for comparative studies aimed at evaluating FastLexRank's performance and efficiency relative to these innovative models.

\section{Ethical Consideration}
In this study, we utilized the Twitter dataset in compliance with the data-sharing agreement, ensuring that no personally identifiable information (PII) is presented. For the four examples listed in Table~\ref{tab:mlRepre}, we removed all PII and ensured that it is impossible to trace the authors based on the provided information.

\bibliography{latex/acl_bib}

\appendix
\section{GPT-4o summary about most representative posts}
\label{sec:appendix}
The collection of tweets emphasizes the importance of participating in the 2020 U.S. Census. Key points highlighted across the tweets include:

\begin{enumerate}
   
\item Impact on Community Resources: Many tweets underscore that census data determines funding for critical services such as schools, roads, public transportation, healthcare, and community resources.

\item Political Representation: Accurate census data is crucial for determining political representation at various levels of government.

\item Misinformation and Participation: Some tweets address the challenges of misinformation and encourage communities to educate themselves to ensure everyone is counted.

\item Civic Duty: Participating in the census is described as a civic duty that affects the allocation of federal funding and public resources for the next decade.

\item Public Awareness Campaigns: Numerous tweets mention efforts by public officials, organizations, and communities to raise awareness and encourage census participation.

\item Inclusivity: The emphasis is on ensuring that everyone, especially hard-to-count populations, is included to achieve a fair and accurate count.

\item Educational Efforts: Various educational resources, videos, and public service announcements are shared to inform people about the census.

\item Partnerships: Tweets highlight partnerships between the U.S. Census Bureau and local organizations, schools, and governments to promote census participation.

\end{enumerate}
Overall, the overarching message is to participate in the 2020 Census to ensure fair distribution of resources, accurate political representation, and better planning for community needs.

\end{document}